\pdfoutput=1

\documentclass[11pt]{article}

\usepackage[final]{acl}

\usepackage{times}
\usepackage{latexsym}
\usepackage{enumitem}
\usepackage[most]{tcolorbox}
\usepackage{authblk}

\newtcolorbox{PromptBox}[1][]{
  colback=gray!5,
  colframe=gray!80!black,
  fonttitle=\bfseries,
  title=Prompt,
  breakable,
  #1
}
\usepackage[T1]{fontenc}

\usepackage[utf8]{inputenc}

\usepackage{microtype}
\usepackage{wrapfig}
\usepackage{inconsolata}

\usepackage{graphicx}
\usepackage{booktabs}
\usepackage{makecell}

\newcommand{\mypar}[1]{\noindent\textbf{#1} \ }

\title{What’s Missing in Vision-Language Models? Probing Their Struggles with Causal Order Reasoning}

\author[1]{\bf Zhaotian Weng}
\author[1]{\bf Haoxuan Li}
\author[3]{\bf Xin Eric Wang}
\author[2]{\bf Kuan-Hao Huang}
\author[1]{\bf Jieyu Zhao}

\affil[1]{University of Southern California}
\affil[2]{Texas A\&M University}
\affil[3]{University of California, Santa Barbara}

\affil[ ]{\texttt{\{wengzhao, lihaoxua, jieyuz\}@usc.edu}, 
\texttt{ericxwang@ucsb.edu},
\texttt{khhuang@tamu.edu}}

\begin{document}
\maketitle
\begin{abstract}
Despite the impressive performance of vision-language models (VLMs) on downstream tasks, their ability to understand and reason about causal relationships in visual inputs remains unclear. Robust causal reasoning is fundamental to solving complex high-level reasoning tasks, yet existing benchmarks often include a mixture of reasoning questions, and VLMs can frequently exploit object recognition and activity identification as shortcuts to arrive at the correct answers, making it challenging to truly assess their causal reasoning abilities. To bridge this gap, we introduce VQA-Causal and VCR-Causal, two new benchmarks specifically designed to isolate and rigorously evaluate VLMs’ causal reasoning abilities. Our findings reveal that while VLMs excel in object and activity recognition, they perform poorly on causal reasoning tasks, often only marginally surpassing random guessing. Further analysis suggests that this limitation stems from a severe lack of causal expressions in widely used training datasets, where causal relationships are rarely explicitly conveyed. We additionally explore fine-tuning strategies with hard negative cases, showing that targeted fine-tuning can improve model's causal reasoning while maintaining generalization and downstream performance. Our study highlights a key gap in current VLMs and lays the groundwork for future work on causal understanding. 
~\footnote{The dataset and code is avaiable at \url{https://github.com/limenlp/CausalVLM}}
\end{abstract}

\section{Introduction}

Pre-trained VLMs have demonstrated impressive performance across a wide range of tasks, including visual question answering~\citep{antol2015vqa,li2019visualbert}, reasoning~\citep{zellers2019recognition}, and object detection~\citep{li2022grounded}. However, strong performance on these benchmarks does not necessarily reflect a rich understanding of visual inputs. Recent studies have revealed that VLMs struggle with tasks demanding high-level visual understanding, such as verb comprehension, spatial reasoning, attribute attachment, and counting~\citep{wang2023paxion,kamath2023s,yuksekgonul2023and,chen2024spatialvlm,parcalabescu-etal-2021-seeing}. Crucially, whether VLMs possess genuine causal reasoning abilities remains largely unexplored. For instance, can VLMs distinguish between ``The woman's holding an umbrella is caused by the rain.'' and ``The rain is caused by the woman holding an umbrella.''? Robust causal understanding and reasoning are fundamental to tackling complex real-world decision making~\citep{lake2017building},
but this capability in VLMs remains largely unexplored.

Existing benchmarks that aim to assess reasoning in VLMs often conflate causal reasoning with other types of reasoning tasks~\citep{antol2015vqa,zellers2019recognition}, and many questions can be answered by object recognition or activity understanding alone. For example, our analysis of the Visual Question Answering (VQA) and Visual Commonsense Reasoning (VCR) benchmarks reveals that only 0.92\% of questions in the VQA validation set~\citep{antol2015vqa} and 35.43\% in the VCR validation set~\citep{zellers2019recognition} involve causal reasoning. Our analysis of 100 randomly selected VCR questions found that 46\% could be answered correctly through object detection or activity understanding alone, without requiring genuine causal reasoning. These issues make it difficult for current benchmarks to independently and effectively evaluate the causal reasoning ability of VLMs.

\begin{figure}[t]
  \centering
  \includegraphics[width=\columnwidth]{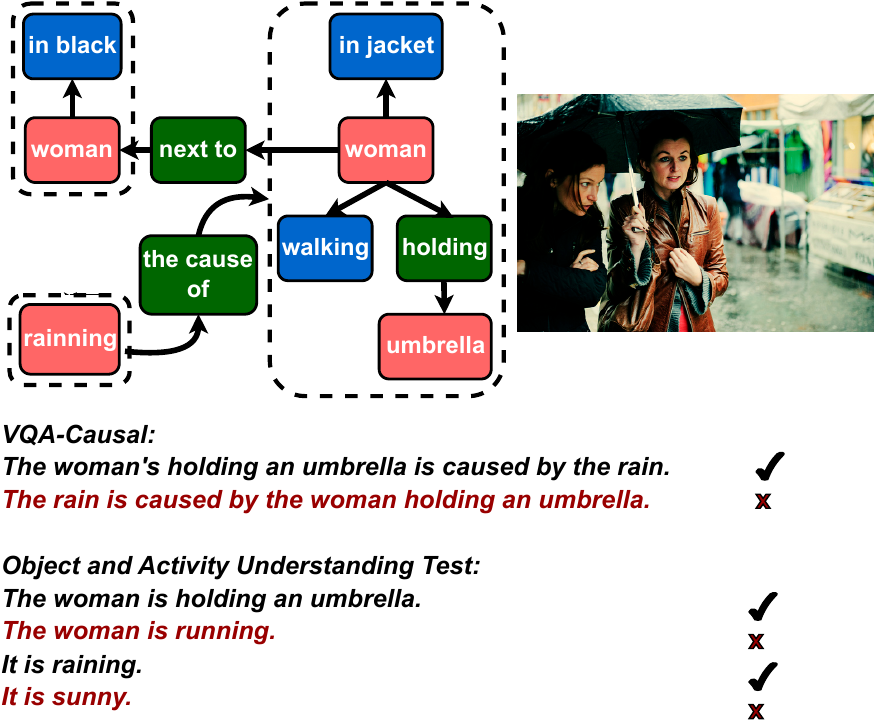}
  \caption{Examples from the VQA-Causal test and the Object and Activity Understanding test. Models tend to focus on low-level visual features such as objects and activities which are represented by the red and blue nodes in the scene graph on the left, but fail to capture high-level visual features such as relationships between activities, especially causal relationships in our case, which are represented by the green nodes in the scene graph. }
  \label{fig:pipeline}
  \vspace{-1.5em}
\end{figure}

To address this gap, we introduce VQA-Causal and VCR-Causal, the first benchmarks specifically designed to rigorously and independently evaluate VLMs’ causal reasoning abilities. Constructed from different sources, VQA~\citep{antol2015vqa} and VCR~\citep{zellers2019recognition}, this dual-benchmark setup enables fine-tuning on VQA data and both in-domain (VQA-Causal) and out-of-domain (VCR-Causal) evaluation, thereby providing a robust assessment of models' causal reasoning capabilities and their generalizability across datasets. VQA-Causal consists of 1,947 instances, and VCR-Causal contains 3,511 instances, with each image paired with 12 caption pairs using different causal conjunctions. Each caption pair differs only in the causal relationship between events, as illustrated in Figure~\ref{fig:pipeline}. This counterfactual approach ensures a comprehensive evaluation of the model's understanding of causal relationships, avoiding potential biases toward specific causal expressions. 

We evaluate 12 widely-used VLMs, covering a broad spectrum of architectures and objectives, including score-based and generative models trained with diverse objectives. The evaluated models include CLIP~\citep{radford2021learning}, NegCLIP~\citep{yuksekgonul2023and}, BLIP~\citep{li2022blip}, FLAVA~\citep{singh2022flava} , LLaVA1.5, LLaVA1.6~\citep{liu2023visual}, Qwen2.5-VL~\citep{bai2025qwen2} and so on. All models perform poorly on both VQA-Causal and VCR-Causal, with ten out of twelve achieving no more than 52\% accuracy, which is only marginally above a random guess (50\%) and significantly below human performance (98\%). 
These findings highlight a fundamental limitation of existing VLMs in causal reasoning.

To better understand whether the poor performance on causal reasoning tasks stems from a lack of basic visual understanding, we constructed a controlled evaluation set by modifying the VQA-Causal dataset. This modified dataset contains the same 1,947 instances as VQA-Causal, but each image is paired with four captions, two of which correctly describe the image. The incorrect captions differ by altering the object or modifying the described activity, as illustrated in Figure~\ref{fig:pipeline}. Our results reveal that while VLMs perform well in recognizing objects and activities, they struggle significantly with reasoning about causal relationships between activities, further reinforcing our findings from VQA-Causal and VCR-Causal.

We then investigate why VLMs trained on large-scale image-text corpora fail to learn causal relationships between events in visual inputs. Focusing on LAION-400M~\citep{schuhmann2021laion} (used by OpenCLIP) and MSCOCO~\citep{lin2014microsoft} (used in FLAVA and NegCLIP)~\citep{singh2022flava,yuksekgonul2023and}, we found that explicit causal expressions are extremely rare. Quantitatively, only 0.08\% of LAION-400M and 0.01\% of MSCOCO instances contain explicit causal expressions. This scarcity explains why VLMs excel at object and activity recognition but struggle with causal reasoning.

To mitigate this limitation, we explored fine-tuning strategies incorporating hard negative cases, captions that differ from the correct ones only in the causal order, demonstrating that targeted fine-tuning can significantly enhance causal reasoning. Our fine-tuned model, CausalCLIP, achieves notable improvements on both in-domain and out-of-domain benchmarks while maintaining strong performance on downstream tasks.

Our contributions are as follows:

\begin{itemize}[topsep=-2pt, itemsep=1pt, leftmargin=18pt]

\item We introduce VQA-Causal and VCR-Causal, the first benchmarks specifically designed to isolate and comprehensively evaluate causal reasoning in VLMs, addressing a critical limitation in existing benchmarks. Moreover, this setup allows us to use one dataset (e.g., VQA) as an in-domain source for fine-tuning and evaluate the model’s causal reasoning ability on both in-domain (VQA-Causal) and out-of-domain (VCR-Causal) benchmarks to assess generalization.

\item Our experimental results reveal that while VLMs excel in object and activity understanding, they perform poorly on causal reasoning tasks, with some only marginally surpassing random guessing. Additionally, our analysis of four widely used datasets for VLM training, fine-tuning, and benchmarking uncovers a severe lack of causal expressions, providing insight into why models fail to learn causal relationships between different activities during training process.
\item We explore fine-tuning with hard negative cases and demonstrate that targeted fine-tuning can enhance causal reasoning performance. Our approach achieves notable improvements on both in-domain and out-of-domain benchmarks while maintaining minimal impact on downstream task performance.
\end{itemize}

\begin{figure}[t]
  \centering
  \includegraphics[width=\columnwidth]{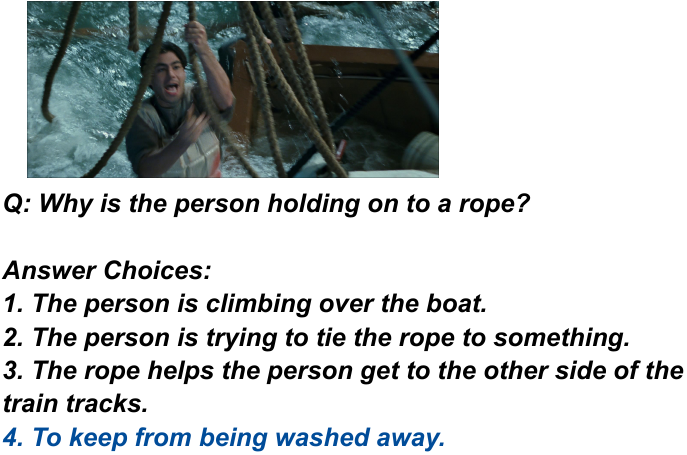}
  \caption{
  The VCR dataset fails to genuinely evaluate a model's causal reasoning ability. In this example, the model can eliminate choice 3 by recognizing that there is no train tracks in the image. It can also rule out captions 2 and 1 by observing that the person is not tying the rope to or climb over anything. As a result, the model can arrive at the correct answer purely through object and activity understanding, without requiring genuine causal reasoning. 
}
  \label{fig:vcr_example}
  \vspace{-1em}
\end{figure}

\section{Benchmarks for Causal Order Reasoning}

Existing benchmarks, such as VQA~\citep{antol2015vqa}, VCR~\citep{zellers2019recognition}, and GQA~\citep{hudson2019gqa}, include questions related to causal reasoning. However, many instances within these datasets involve multiple types of reasoning, making it difficult to isolate and evaluate a model's specific understanding of causal relationships. Additionally, a significant portion of the causal reasoning examples fail to truly assess a model’s comprehension of causality.

For example, as illustrated in Figure~\ref{fig:vcr_example}, the VCR~\citep{zellers2019recognition} question ``Why is the person holding on to a rope?'' allows a model to select the correct answer by merely identifying the absence of specific objects, such as train tracks, in the image, thus eliminating an incorrect option. Furthermore, by recognizing that the depicted activity is not ``tying the rope to something,'' or ``climbing over the boat'' the model can exclude another 2 options. With only basic object and activity recognition, the model can reach the correct answer without demonstrating genuine reasoning about the causal relationships between different entities in the visual input.

In contrast, our proposed datasets place a strong emphasis on requiring models to understand the causal relationships between various events within the visual input.
Our newly developed evaluation corpora adopt the format proposed by ~\citet{kamath2023s}, featuring an image paired with several captions that vary only in causal order. Specifically, VQA-Causal and VCR-Causal are constructed from the widely-used VQA and VCR datasets~\citep{antol2015vqa,zellers2019recognition}. A key contribution of our work is that every instance in our dataset demands models to genuinely reason about the causal relationships between different events in the visual input, rather than taking shortcuts by merely identifying objects and activities to arrive at the correct answer.

\subsection{Dataset Construction}
\mypar{VQA-Causal}
We constructed the VQA-Causal dataset using the validation set and validation annotation files from the VQA dataset~\citep{antol2015vqa}. Specifically, we selected all instances from the validation set where the questions contained the keyword ``Why'' to form our VQA-Causal dataset. Each original question and answer pair was transformed into two sentences connected by causal conjunctions, differing only in the causal order while keeping everything else identical. For example, given an image with the original VQA question ``Why is the woman holding an umbrella?'' and the correct answer ``It is raining'', we retained the image and generated two captions using the causal conjunction \textit{is caused by }: 
``The woman's holding an umbrella is caused by the rain.'' and ``The rain is caused by the woman holding an umbrella.'', as illustrated in Figure~\ref{fig:pipeline}.

We used 12 causal conjunctions to create 12 groups of caption pairs for each image. These conjunctions were carefully chosen to capture variations in the syntactic ordering of causes and effects, as such variations can potentially influence a model’s performance on causal reasoning tasks. Specifically, some conjunctions, such as \textit{is due to}, \textit{is caused by}, \textit{is a result of}, \textit{is the effect of}, \textit{is the consequence of}, \textit{because}, and \textit{owe to}, place the effect before the cause in the sentence structure. In contrast, others such as \textit{result in}, \textit{cause}, \textit{lead to}, \textit{give rise to}, and \textit{bring about to}, place the cause before the effect. Each group contains one caption expressing a correct causal relationship and one expressing an incorrect relationship, differing only in the causal direction.

In total, we extracted 1,947 instances from the VQA dataset, with each image paired with 12 distinct caption pairs. This setup offers several advantages: (1) It enables a rigorous evaluation of the model’s ability to reason about causal relationships within visual inputs. (2) The use of diverse causal conjunctions allows us to assess the model’s understanding and sensitivity to different causal expressions, while also mitigating potential biases that may arise from over-reliance on any single conjunction during the reasoning process.

\mypar{VCR-Causal}
Similarly, we constructed the VCR-Causal dataset using the validation set and annotation files from the VCR dataset. We selected instances containing ``Why'' in their questions to form the VCR-Causal dataset. The VCR-Causal dataset contains a total of 3,511 instances, with each image associated with 12 caption pairs.

We conducted human verification on a randomly sampled subset of both VQA-Causal and VCR-Causal. Two human annotators with NLP backgrounds were asked to judge whether captions for each instance were (1) semantically coherent given the image context and (2) fluent. Each annotator reviewed 200 image-caption pairs from each dataset. Results show that over 96\% of the captions were rated as both fluent and reasonable, indicating that our generation process yields high-quality, interpretable inputs for evaluating causal reasoning.

\subsection{Causal Order Reasoning Test}
\mypar{Task}  
For the VQA-Causal and VCR-Causal benchmarks, we follow the experimental setup used by ~\citet{kamath2023s}. The input consists of an image paired with two caption options, which differ only in their causal order, as illustrated in Figure~\ref{fig:pipeline}. Consistent with ~\citet{kamath2023s}, we evaluate the models under a zero-shot setting. Our evaluation metric is the proportion of images for which the matching score between the image and the correct caption is higher than the matching score between the image and the incorrect caption.  

\mypar{Models}  
We select both score-based and text-generation based models: 

\begin{itemize}
    \item \textbf{Score-based models:} 
    CLIP ViT-B/32, CLIP ViT-L/14~\citep{radford2021learning}, FLAVA~\citep{singh2022flava}, BLIP ITM ViT-B, BLIP ITM ViT-L~\citep{li2022blip}, BLIP2 ITM~\citep{li2023blip}, BLIP2 Feature Extractor~\citep{li2023blip}, NegCLIP~\citep{yuksekgonul2023and}, and RobustCLIP~\citep{schlarmann2024robust}. These models produce matching scores for each image-caption pair independently. Among them, NegCLIP is fine-tuned with hard negatives samples, making it more sensitive to the word order and RobustCLIP is fine-tuned with adversarial augmentations to improve the model’s robustness.
    \item \textbf{Text-generation models:} LLaVA1.5-7B, LLaVA1.6-7B~\citep{liu2023visual}, Vicuna1.5-7B~\citep{vicuna2023,zheng2023judging}, Qwen2.5-VL-7B~\citep{bai2025qwen2},Qwen3-VL-8B~\citep{Qwen3-VL}.
    We include Vicuna to validate that a language model relying solely on text input cannot effectively solve the causal reasoning tasks in our benchmark, thereby demonstrating the benchmark's reliability. By comparing Vicuna with LLaVA, which takes both image and text inputs, we further investigate whether LLaVA is capable of leveraging visual information to support causal reasoning.
  
\end{itemize}

For LLaVA and Qwen2.5-VL, we follow the settings in ~\citet{kamath2023s} and reformulate the task by converting the two captions into two questions. For example:  
\begin{enumerate}[topsep=-2pt, itemsep=1pt, leftmargin=18pt]
    \item ``The woman's holding an umbrella is caused by the rain. Does it reflect the proper causal relationship?''
    \item ``The rain is caused by the woman holding an umbrella. Does it reflect the proper causal relationship?''
\end{enumerate}  
\vspace{0.5em}
We measure the probabilities of models answering ``yes'' or ``no'' to these questions. Correctness is determined based on one of the following criteria:  
\begin{enumerate}[topsep=-2pt, itemsep=1pt, leftmargin=18pt]
    \item The probability of ``yes'' is highest for the gold option.~\citep{kamath2023s}
    \item If both answers are ``no'', the probability of ``no'' is the lowest for the gold option.
\end{enumerate}  

\subsection{Benchmarking Results}
Table~\ref{tab:benchmarkresults} present the performance of nine score-based models and four generation-based models on our VQA-Causal and VCR-Causal benchmarks, respectively. Overall, all models perform near random and far below human estimate, revealing a clear lack of robust causal reasoning ability in current VLMs. Detailed results for each model across the twelve causal conjunctions are provided in Table~\ref{tab:vqacausalresults} and Table~\ref{tab:vcrcausalresults}.

\begin{table}[htbp]
    \centering
    \normalsize
    \setlength{\tabcolsep}{4pt}
    \begin{tabular}{l|c|c}
        \toprule
        Model & \textbf{VQA-Causal} & \textbf{VCR-Causal} \\
        \midrule
        \multicolumn{1}{c}{} & \multicolumn{2}{c}{Score-Based Models} \\
        \midrule
        BLIP ITM Base & 48.94 & 50.66 \\
        BLIP ITM Large & 48.68 & 47.99 \\
        BLIP2 ITM & 50.76 & 49.95 \\
        BLIP2 FE & 51.51 & 50.76 \\
        CLIP ViT B/32 & 51.62 & 50.35 \\
        CLIP ViT L/14 & 50.74 & 51.66 \\
        NegCLIP & 50.89 & 51.30 \\
        RobustCLIP & 50.66 & 53.68 \\
        FLAVA & 48.52 & 49.99 \\
        \midrule
        \multicolumn{1}{c}{} & \multicolumn{2}{c}{Text-Generation Models} \\
        \midrule
        Vicuna 1.5 & 50.86 & 56.03 \\
        LLaVA 1.5 & 53.19 & 52.12 \\
        LLaVA 1.6 & 58.11 & 57.24   \\
        Qwen2.5-VL & 50.66 & 47.26 \\
        Qwen3-VL & 42.01 & 43.73 \\
        \midrule
        Random & 50.00 & 50.00 \\
        Human Estimate & 99.17 & 98.17 \\
        \bottomrule
    \end{tabular}
    \caption{Accuracy on the causal‐order reasoning tests of 12 VLMs for the VQA-Causal and VCR-Causal benchmarks. All models perform only marginally above random guessing and remain significantly below human-level performance. ``BLIP2 FE'' denotes the BLIP2 feature extractor model. Detailed results for each of the twelve causal conjunctions are provided in Table~\ref{tab:vqacausalresults} and Table~\ref{tab:vcrcausalresults} in Appendix.}
    \label{tab:benchmarkresults}
\end{table}

\mypar{Causal Conjunctions Performance} 
Across both the VQA-Causal and VCR-Causal benchmarks, the CLIP model family, including CLIP ViT B/32, CLIP ViT L/14, NegCLIP and RobustCLIP~\citep{radford2021learning,yuksekgonul2023and,schlarmann2024robust}, demonstrates relatively stronger performance on conjunctions such as is caused by, is due to, is the consequence of, because, owe to, and is the effect of. These conjunctions share a common syntactic structure in which the result precedes the cause. In contrast, these models perform notably worse on conjunctions such as result in, cause, lead to, give rise to, and bring about to, where the cause appears before the result. However, FLAVA~\citep{singh2022flava} exhibits the opposite trend. On the VQA-Causal benchmark, it performs relatively poorly on conjunctions where the result comes first, but shows stronger performance on those where the cause precedes the result. These observations suggest that the syntactic ordering of cause and effect within a sentence plays a critical role in model performance, and that certain models may be sensitive to specific linguistic patterns of causal expression.

\mypar{Impact of Prior Fine-Tuning Strategies} Fine-tuning for caption order improves a model's sensitivity to word order, thereby improving its performance on certain causal order tests. For instance, NegCLIP outperforms CLIP models when tested on conjunctions like is due to and is caused by in most cases, demonstrating substantial improvements. However, for conjunctions like result in, cause, and lead to, NegCLIP underperforms compared to CLIP models. This suggests that fine-tuning for word order amplifies the model's strengths for specific conjunctions but also exacerbates its weaknesses for others, particularly those it initially struggled with. Moreover, adversarial robustness fine-tuning, as implemented in RobustCLIP, does not lead to significant improvements in causal order reasoning performance.

\section{Activity and Object Understanding Test}
To further investigate whether the poor causal reasoning performance of VLMs arises from a lack of understanding of entities in visual inputs, we conducted the Activity and Object Understanding Test. The results show that VLMs exhibit strong capabilities in recognizing objects and activities within images. This suggests that VLMs tend to focus on learning low-level visual features such as objects and activities recognition but fail to capture high-level features like causal relationships between activities.

\mypar{Data Construction}
We extended the VQA-Causal dataset to construct this evaluation. For each original instance, we generated four captions: two correct captions that preserve the original causal event but decompose it into independent factual statements, and two incorrect captions, which were carefully crafted by modifying the object or the activity from the correct captions to make them factually inaccurate. This setup allows us to isolate the model’s understanding of objects and activities from its ability to reason about causal relationships. An illustration is provided in Figure~\ref{fig:pipeline}.

\mypar{Object and Activity Understanding Test}
We conducted experiments with all score-based VLMs mentioned in Section 2.2 
to assess their understanding of objects and activities within the input images. The input to each model consisted of an image paired with four captions described in the last paragraph, as illustrated in Figure~\ref{fig:pipeline}. 
We considered the model's response correct if the two captions with the highest scores were the correct ones. This task setup requires models to accurately understand both the objects and activities depicted in the input image to achieve a correct response.

\mypar{Results and Analysis}
As shown in Table~\ref{tab:OAtest}, all models achieve strong performance on the object and activity understanding task. In Figure~\ref{fig:pipeline}, the red nodes represent objects, the blue nodes indicate the attributes of these objects, and the green nodes depict the relationships between different objects and activities. VLMs tend to focus on learning low-level features which is the red and blue node in the scene graph but fail to capture high-level features which is the green node representing the relationships between objects and activities. This limitation in capturing structured visual relationships may explain why VLMs perform close to random on high-level reasoning tasks, including causal reasoning, as well as on other complex reasoning tasks like spatial reasoning~\citep{kamath2023s} and verb understanding~\citep{wang2023paxion}, which have been highlighted in previous studies.

\begin{table}[htbp]
  \centering
  \normalsize
  \setlength{\tabcolsep}{6pt}  
  \begin{tabular}{l|cc}        
    \toprule
    \textbf{Model}     & \textbf{VQA-Causal} & \textbf{O\&A Test} \\
    \midrule
    BLIP ITM Base      & 48.94        & 94.61              \\
    BLIP ITM Large     & 48.68        & 95.53              \\
    BLIP2 ITM          & 50.76        & 92.24              \\
    BLIP2 FE & 51.51        & 83.98              \\
    CLIP ViT B/32      & 51.62        & 76.53              \\
    CLIP ViT L/14      & 50.74        & 85.31             \\
    NegCLIP            & 50.89       & 87.62             \\
    RobustCLIP         & 50.66       & 83.26              \\
    FLAVA              & 48.52       & 71.85            \\
    \bottomrule
  \end{tabular}
  \caption{Accuracy on the Object and Activity Understanding Test (\textit{O\&A Test}) versus the Causal Order Reasoning Test (\textit{VQA-Causal}). All models exhibit strong performance on the \textit{O\&A Test}, indicating that while VLMs effectively recognize objects and activities, they struggle with causal reasoning task. ``BLIP2 FE'' denotes the BLIP2 feature extractor model.}
  \label{tab:OAtest}
\end{table}

\section{Why Struggling with Causal Reasoning? A Data-Level Exploration}

All evaluated VLMs were pretrained and fine-tuned on large-scale image-text corpora and have shown strong performance on traditional benchmarks. To explore why they fail to learn causal relationships, we investigate this limitation from a data-level perspective.

We selected four widely-used datasets for VLM pretraining and benchmark: LAION-400M~\citep{schuhmann2021laion}, which was used to train OpenCLIP~\citep{ilharco_gabriel_2021_5143773,cherti2023reproducible,radford2021learning,schuhmann2022laionb}, and MSCOCO~\citep{lin2014microsoft}, which was used in FLAVA's training and NegCLIP's fine-tuning. For benchmark datasets, we analyzed VQA and VCR~\citep{antol2015vqa,zellers2019recognition}, two standard datasets commonly used to evaluate the reasoning capabilities of VLMs.

\mypar{Pre-Training Datasets} We randomly sampled about 5,200,000 captions from the LAION-400M dataset to examine the prevalence of causal expressions. Specifically, we looked for captions containing any of the following causal-related terms: \textit{because}, \textit{cause}, \textit{lead to}, \textit{reason}, \textit{is the reason why}, \textit{is the effect of}, \textit{owe to}, \textit{give rise to}, \textit{bring about to}, \textit{result in}. Among the sampled captions, only 4,026 captions ($\sim$0.08\%) included causal expressions. Similarly, in the MSCOCO dataset, where we analyzed 415,795 captions, only 53 captions ($\sim$0.01\%) contained causal relationships.

These findings reveal that causal relationships are exceedingly rare in the training datasets, with less than 0.1\% of captions involving causal reasoning, making it difficult for VLMs to learn and generalize causal understanding from visual inputs.

\begin{table*}[!t]
    \centering
    \vspace{0.5em}
    \normalsize
    \setlength{\tabcolsep}{5pt}
    \resizebox{\textwidth}{!}{
    \begin{tabular}{l|c|cccccccccccc}
        \toprule
        & \multicolumn{13}{c}{\textbf{VQA-Causal (In-Domain)}} \\
        Model & Avg & CW1 & CW2 & CW3 & CW4 & CW5 & CW6 & CW7 & CW8 & CW9 & CW10 & CW11 & CW12 \\
        \midrule
        NegCLIP & 50.89 & 62.51 & 62.25 & 64.77 & 67.33 & 66.97 & 59.94 & 69.75 & 28.45 & 39.39 & 33.85 & 29.33 & 26.14 \\
        CausalCLIP & 61.46 & 70.57 & 71.70 & 69.54 & 73.81 & 72.16 & 61.43 & 56.45 & 55.68 & 48.84 & 57.22 & 53.72 & 46.38 \\
        \midrule
        & \multicolumn{13}{c}{\textbf{VCR-Causal (Zero-Shot)}} \\
        Model & Avg & CW1 & CW2 & CW3 & CW4 & CW5 & CW6 & CW7 & CW8 & CW9 & CW10 & CW11 & CW12 \\
        \midrule
        NegCLIP       & 51.30 & 52.21 & 53.89 & 54.80 & 55.23 & 57.11 & 50.04 & 50.87 & 47.22 & 50.19 & 51.32 & 45.43 & 47.34 \\
        CausalCLIP & 57.37 & 59.10 & 62.49 & 61.29 & 63.69 & 62.92 & 54.97 & 50.16 & 58.96 & 53.00 & 58.33 & 51.38 & 52.15 \\
        \bottomrule
    \end{tabular}
    }
    \caption{
    CausalCLIP demonstrates strong generalization on both VQA-Causal (in-domain) and VCR-Causal (zero-shot) benchmarks. CW1--CW12 correspond to the following twelve causal conjunctions: \textit{is due to}, \textit{is caused by}, \textit{is a result of}, \textit{is the effect of}, \textit{is the consequence of}, \textit{because}, \textit{owe to}, \textit{result in}, \textit{cause}, \textit{lead to}, \textit{give rise to}, and \textit{bring about to}.
    }
    \label{tab:finetuneresults}
    \vspace{-1em}
\end{table*}

\begin{table*}[!t]
    \centering
    \vspace{0.5em}
    \normalsize
    \setlength{\tabcolsep}{4pt}
    \resizebox{\textwidth}{!}{
    \begin{tabular}{l|cccccccc|cccccccc}
        \toprule
        \multicolumn{1}{c|}{} &
        \multicolumn{8}{c|}{\textbf{COCO}} &
        \multicolumn{8}{c}{\textbf{Flickr30K}} \\
        
        Model & 
        
        \makecell{M1} & \makecell{M2} & \makecell{M3} & \makecell{M4} & 
        \makecell{M5} & \makecell{M6} & \makecell{M7} & \makecell{M8} &
        \makecell{M1} & \makecell{M2} & \makecell{M3} & \makecell{M4} & 
        \makecell{M5} & \makecell{M6} & \makecell{M7} & \makecell{M8} \\
        \midrule
        OpenCLIP   & 0.30 & 0.56 & 0.50 & 0.75 & 0.30 & 0.56 & 0.10 & 0.34 & 0.59 & 0.84 & 0.79 & 0.95 & 0.59 & 0.84 & 0.16 & 0.57 \\
        NegCLIP    & 0.41 & 0.68 & 0.56 & 0.80 & 0.41 & 0.68 & 0.11 & 0.39 & 0.67 & 0.89 & 0.79 & 0.95 & 0.67 & 0.89 & 0.16 & 0.62 \\
        CausalCLIP & 0.38 & 0.64 & 0.54 & 0.78 & 0.38 & 0.64 & 0.11 & 0.38 & 0.64 & 0.87 & 0.78 & 0.94 & 0.64 & 0.87 & 0.16 & 0.60 \\
        \bottomrule
    \end{tabular}
    }
    \caption{CausalCLIP exhibits minimal performance loss compared to NegCLIP and even outperforms OpenCLIP on retrieval tasks across both MSCOCO and Flickr30K datasets. Metrics M1–M8 correspond to: ImagePrec@1, ImagePrec@5, TextPrec@1, TextPrec@5, ImageRecall@1, ImageRecall@5, TextRecall@1, and TextRecall@5, respectively.}
    \label{tab:downstream}
    \vspace{-1em}
\end{table*}

\mypar{Benchmark Datasets} We then examined the causal reasoning content of two commonly used VLM benchmarks: VQA and VCR~\citep{antol2015vqa,zellers2019recognition}. In the VQA validation set, only 1,962 out of 214,354 questions ($\sim$0.92\%) involved causal reasoning related to visual inputs; In the VCR validation set, 9,401 out of 26,534 questions ($\sim$35.43\%) involved causal reasoning.

To further analyze the VCR dataset, we randomly selected 100 questions from the subset involving causal reasoning and conducted a detailed human annotation. We found that 46\% of these questions could be answered correctly by relying solely on object detection or activity understanding without requiring any genuine understanding of causal relationships. As shown in Figure~\ref{fig:vcr_example}, a model could eliminate one incorrect option by recognizing the absence of objects such as ``train tracks'' in the image. Furthermore, the model could identify the actions of the person in the image (e.g., not tying a rope to something or climbing over the boat) to select the correct answer. In such cases, models rely on object detection and activity recognition to arrive at the correct answer without reasoning about the causal relationships between events in the image and thus have good performance on such benchmarks.

\section{Data-Level Improvement}

We extract a subset of data from the VQA~\citep{antol2015vqa} training set and for each instance generate 10 caption pairs, each consisting of one correct caption reflecting a valid causal relationship and one incorrect caption as a hard negative example. These examples are used to fine-tune the models. This fine-tuning strategy significantly improves the models' causal reasoning performance on both in-domain and out-of-domain datasets, while preserving downstream performance.

\mypar{Dataset}
Following the VQA-Causal construction methodology, we extracted all ``why'' questions from the VQA training set along with their corresponding correct answers, resulting in a total of 4,891 instances. For each instance, we constructed 10 caption pairs using ten different causal conjunctions: \textit{is due to}, \textit{is caused by}, \textit{is a result of}, \textit{is the effect of}, \textit{because}, \textit{result in}, \textit{cause}, \textit{lead to}, \textit{give rise to}, and \textit{bring about to}. Each pair consists of two captions that differ only in the direction of the causal relationship, with all other elements remaining identical.

\mypar{Finetuning} We adopt the fine-tuning setup from NegCLIP~\citep{yuksekgonul2023and} and extend CLIP's~\citep{radford2021learning} contrastive learning objective to better support causal reasoning. For each image-caption pair, we introduce hard negative captions, including (1) the incorrect causal order caption for the same image, and (2) three randomly sampled negative captions from other instances in the dataset. Additionally, we randomly sample one alternative image per instance to serve as a negative image, helping ensure generalization and reduce overfitting.

We conduct fine-tuning experiments using NegCLIP~\citep{yuksekgonul2023and}, a ViT-B/32 variant of CLIP. For each batch of \(N\) images \(I_N\), we concatenate the \(N\) corresponding correct captions and \(N\) incorrect captions to form a \(2N\) caption batch. We then compute a similarity matrix between all images and all captions. Following ~\citet{yuksekgonul2023and}, we obtain both row-wise and column-wise cross-entropy losses, while ignoring the loss terms from negative captions in the column-wise direction.

\mypar{Baseline} Since we fine-tune on the NegCLIP model, its original performance on the causal reasoning benchmarks serves as our baseline. It is worth noting that our fine-tuning only uses 10 causal conjunctions and is performed exclusively on data from the VQA training set. However, we evaluate the model on all 12 causal conjunctions using both VQA-Causal (as the in-domain benchmark) and VCR-Causal (as the out-of-domain benchmark). Notably, VCR-Causal serves as a zero-shot test set. This setup allows us to evaluate the model's generalization in two ways: (1) to unseen causal conjunctions not present during fine-tuning, and (2) to out-of-domain dataset, thereby providing a more comprehensive assessment of its causal reasoning abilities.

\mypar{Evaluation} As shown in Table~\ref{tab:finetuneresults}, our fine-tuned model \textit{CausalCLIP} achieves strong causal reasoning performance on both in-domain and out-of-domain benchmarks. Furthermore, Table~\ref{tab:downstream} shows that this fine-tuning strategy preserves downstream performance and even outperforms OpenCLIP on retrieval tasks over MSCOCO~\citep{lin2014microsoft} and Flickr30k~\citep{young2014image}, following the setup of~\citet{yuksekgonul2023and}.

\section{Related Work}
VLMs have excelled across a wide range of multimodal tasks, including object detection~\citep{li2022grounded,zhang2022glipv2}, image-text retrieval ~\citep{radford2021learning,li2022blip,li2023blip} , visual question answering~\citep{li2019visualbert,antol2015vqa,liu2023visual} and commonsense reasoning~\citep{zellers2019recognition}. However, recent benchmarks have revealed that VLMs perform poorly on tasks requiring fine-grained reasoning skills, such as counting~\citep{parcalabescu-etal-2021-seeing,paiss2023teaching}, spatial reasoning~\citep{kamath2023s,cheng2024spatialrgpt,wang2024picture}, verb understanding~\citep{wang2023paxion,hendricks-nematzadeh-2021-probing}, attribute composition\citep{tang2023lemons,weng-etal-2024-images,zhao2022vl,yuksekgonul2023and,weng2025situationalpriv}. These suggest that models fail to possess high-level visual understanding beyond low-level recognition.

Among these reasoning abilities, causal reasoning is one of the most foundamental abilities, as it allows models to plan interventions and infer underlying mechanisms crucial for complex real-world decision making tasks~\citep{lake2017building}, but remains largely underexplored. Many existing benchmarks for evaluating reasoning ability often focus on video-language models~\citep{parmar2024causalchaos,rawal2024dissecting,li2022representation,xiao2021next} and frequently conflate causal reasoning with other forms of reasoning~\citep{yin2021broaden,lu2018r,ben2019vqa,li2018vqa, antol2015vqa,zellers2019recognition,hudson2019gqa,marino2019ok}. Moreover, ~\citet{li2025multimodal} introduced MuCR by generating images conditioned on causes and designing tasks such as selecting the correct effect from multiple candidates. However, many similar benchmarks can still be solved through shortcut strategies like by detecting salient objects or identifying specific activities, as illustrated in Figure~\ref{fig:vcr_example}. This makes it difficult to determine whether models truly capture the causal order between causes and effects.

Our work addresses this critical gap by introducing VQA-Causal and VCR-Causal, that explicitly evaluate whether VLMs can distinguish between alternative causal interpretations of the same visual scene, thus enabling rigorous causal reasoning evaluation in multimodal models.

\section{Conclusion}
We introduce VQA-Causal and VCR-Causal, the first benchmarks for systematically evaluating VLMs' causal reasoning across 12 causal conjunctions. Despite strong performance in object and activity recognition, all twelve evaluated models struggle, with ten scoring no more than 52\% accuracy, barely above chance. To explain this limitation, we analyze four widely used datasets, LAION-400M, MSCOCO, VQA, and VCR, and find explicit causal expressions exceedingly rare in LAION-400M and MSCOCO datasets, with fewer than 0.1\% of instances involving causal relationships. Moreover, only 0.92\% of VQA and 35.43\% of VCR questions require causal reasoning, and 46\% of sampled VCR questions can be solved using shortcuts without genuine causal reasoning.
Finally, we extract 4,891 causality-related instances from the VQA training set and construct contrastive training data by pairing correct captions with hard negative examples that differ only in causal direction. Fine-tuning with this data significantly improves causal reasoning performance on both in-domain and out-of-domain benchmarks, while maintaining downstream task performance. 

\section{Limitations}
Our work uncovers the weaknesses of current VLMs on causal reasoning tasks. By analyzing both their training data and benchmark datasets, we proposed a data-level fine-tuning strategy that significantly enhances causal reasoning ability with minimal impact on downstream performance. However, this approach mainly focuns on data-level and does not address the underlying model architecture. A promising direction for future research is to improve causal and fine-grained reasoning at the model architectural level. For example, researchers could adjust attention weights to guide the model foucs more on fine-grained visual features or implement broadly generalizable modifications to specific VLM components to improve both causal reasoning ability and fine-grained visual understanding. Finally, although our study focuses on vision–language models, causal reasoning and fine-grained visual reasoning in other multimodal settings, such as video–language models, remains an important direction for further investigation.  

\section*{Acknowledgments}
This work was supported by the National Science Foundation through the NAIRR Pilot under award NSF NAIRR250198.

\bibliography{custom}

\appendix

\section{Appendix}
\label{sec:appendix}

\begin{table*}[!t]
    \centering
    \normalsize
    \setlength{\tabcolsep}{4pt}
    \resizebox{\textwidth}{!}{
    \begin{tabular}{l|c|cccccccccccc}
        \toprule
        & \multicolumn{13}{c}{\textbf{VQA-Causal}} \\
        Model & Avg & \makecell{CW1} & \makecell{CW2} & \makecell{CW3} & \makecell{CW4} & \makecell{CW5} & CW6 & \makecell{CW7} & \makecell{CW8} & CW9 & \makecell{CW10} & \makecell{CW11} & \makecell{CW12} \\
        \midrule
        \multicolumn{1}{c}{} & \multicolumn{13}{c}{Score-Based Models} \\
        \midrule
        BLIP ITM Base & 48.94 & 49.36 & 51.52 & 47.92 & 48.54 & 51.16 & 50.90 & 52.23 & 47.87 & 47.35 & 46.84 & 47.05 & 46.53 \\
        BLIP ITM Large & 48.68 & 44.84 & 48.74 & 46.12 & 45.15 & 49.15 & 49.97 & 49.00 & 50.95 & 49.82 & 50.64 & 50.44 & 49.36 \\
        BLIP2 ITM & 50.76 & 59.89 & 61.84 & 58.55 & 59.37 & 57.68 & 59.78 & 61.43 & 36.52 & 40.63 & 37.75 & 37.90 & 37.80 \\
        BLIP2 FE & 51.51 & 57.32 & 60.71 & 58.91 & 59.89 & 58.60 & 58.45 & 58.65 & 40.16 & 40.88 & 40.83 & 42.27 & 41.45 \\
        CLIP ViT B/32 & 51.62 & 57.16 & 57.06 & 58.19 & 58.76 & 61.48 & 56.24 & 58.14 & 42.94 & 43.97 & 43.25 & 42.73 & 39.50 \\
        CLIP ViT L/14 & 50.74 & 59.48 & 59.22 & 54.60 & 56.14 & 57.06 & 60.04 & 58.19 & 40.11 & 41.81 & 42.06 & 39.24 & 40.93 \\
        NegCLIP & 50.89 & 62.51 & 62.25 & 64.77 & 67.33 & 66.97 & 59.94 & 69.75 & 28.45 & 39.39 & 33.85 & 29.33 & 26.14 \\
        RobustCLIP & 50.66 & 58.29 & 58.91 & 54.29 & 54.90 & 56.19 & 59.37 & 58.24 & 41.91 & 42.94 & 43.09 & 39.65 & 40.16 \\
        FLAVA & 48.52 & 40.06 & 40.37 & 39.50 & 44.12 & 42.06 & 38.06 & 41.91 & 56.50 & 61.33 & 60.25 & 60.04 & 58.09 \\
        \midrule
        \multicolumn{1}{c}{} & \multicolumn{13}{c}{Text-Generation Models} \\
        \midrule
        Vicuna 1.5 & 50.86 & 44.94 & 49.56 & 50.54 & 46.07 & 53.21 & 37.24 & 38.16 & 58.96 & 58.35 & 56.60 & 61.12 & 55.52 \\
        LLaVA 1.5 & 53.19 & 57.83 & 58.45 & 55.11 & 61.89 & 54.24 & 53.06 & 55.01 & 50.80 & 52.59 & 44.53 & 44.94 & 49.87 \\
        LLaVA 1.6 & 58.11 & 67.13 & 67.28 & 67.08 & 65.59 & 69.80 & 64.41 & 64.00 & 46.28 & 47.46 & 44.38 & 48.07 & 45.81 \\
        Qwen2.5-VL & 50.66 & 55.37 & 57.11 & 55.26 & 56.81 & 58.86 & 55.93 & 56.45 & 48.02 & 35.18 & 44.27 & 43.97 & 40.73 \\
        Qwen3-VL & 42.01 & 48.74 & 51.31 & 50.59 & 51.87 & 51.87 & 46.69 & 48.69 & 37.70 & 25.12 & 31.23 & 31.84 & 28.81 \\
        \midrule
        Random & 50.00 & 50.00 & 50.00 & 50.00 & 50.00 & 50.00 & 50.00 & 50.00 & 50.00 & 50.00 & 50.00 & 50.00 & 50.00 \\
        Human Estimate & 99.17 & 100.00 & 98.00 & 98.00 & 98.00 & 98.00 & 100.00 & 98.00 & 100.00 & 100.00 & 100.00 & 100.00 & 100.00 \\
        \bottomrule
    \end{tabular}
    }
    \caption{Performance of ten VLMs on the VQA-Causal benchmark. All models perform only slightly better than random guess and remain significantly below human-level performance. CW1--CW12 correspond to the following twelve causal conjunctions: \textit{is due to}, \textit{is caused by}, \textit{is a result of}, \textit{is the effect of}, \textit{is the consequence of}, \textit{because}, \textit{owe to}, \textit{result in}, \textit{cause}, \textit{lead to}, \textit{give rise to}, and \textit{bring about to}. ``BLIP2 FE'' denotes the BLIP2 feature extractor model.}
    \label{tab:vqacausalresults}
\end{table*}

\begin{table*}[ht]
    \centering
    \vspace{0.5em}
    \normalsize
    \setlength{\tabcolsep}{4pt}
    \resizebox{\textwidth}{!}{
    \begin{tabular}{l|c|cccccccccccc}
        \toprule
        & \multicolumn{13}{c}{\textbf{VCR-Causal}} \\
        Model & Avg & CW1 & CW2 & CW3 & CW4 & CW5 & CW6 & CW7 & CW8 & CW9 & CW10 & CW11 & CW12 \\
        \midrule
        \multicolumn{1}{c}{} & \multicolumn{13}{c}{Score-Based Models} \\
        \midrule
        BLIP ITM Base         & 50.66 & 45.60 & 49.81 & 45.17 & 45.88 & 45.71 & 44.43 & 46.08 & 55.51 & 53.72 & 61.35 & 58.53 & 56.14 \\
        BLIP ITM Large        & 47.99 & 38.51 & 42.30 & 39.48 & 36.40 & 40.90 & 44.15 & 40.67 & 60.61 & 53.77 & 60.64 & 60.61 & 57.82 \\
        BLIP2 ITM              & 49.95 & 40.24 & 41.98 & 51.87 & 46.57 & 42.47 & 50.47 & 39.93 & 57.85 & 55.03 & 50.41 & 62.32 & 60.27 \\
        BLIP2 FE & 50.76 & 40.56 & 49.05 & 51.47 & 48.05 & 49.50 & 44.32 & 40.67 & 54.71 & 60.21 & 55.00 & 56.48 & 59.10 \\
        CLIP ViT B/32 & 50.35 & 55.51 & 48.05 & 51.64 & 49.16 & 52.49 & 54.34 & 46.45 & 49.19 & 46.57 & 52.38 & 50.41 & 47.99 \\
        CLIP ViT L/14 & 51.66 & 52.72 & 49.30 & 55.51 & 53.00 & 52.58 & 53.72 & 50.04 & 53.00 & 45.66 & 55.40 & 49.79 & 49.25 \\
        NegCLIP       & 51.30 & 52.21 & 53.89 & 54.80 & 55.23 & 57.11 & 50.04 & 50.87 & 47.22 & 50.19 & 51.32 & 45.43 & 47.34 \\
        RobustCLIP     & 53.68 & 62.29 & 60.35 & 61.24 & 60.89 & 60.75 & 63.94 & 55.11 & 46.28 & 39.33 & 48.50 & 42.75 & 42.67 \\
        FLAVA & 49.99 & 49.42 & 49.90 & 52.18 & 49.47 & 53.32 & 46.48 & 53.00 & 46.23 & 49.36 & 48.11 & 50.53 & 51.89\\
        \midrule
        \multicolumn{1}{c}{} & \multicolumn{13}{c}{Text-Generation Models} \\
        \midrule
        Vicuna 1.5 & 56.03 & 48.48 & 53.83 & 59.30 & 49.96 & 62.77 & 42.04 & 37.94 & 66.93 & 61.72 & 61.01 & 66.76 & 61.63 \\
        LLaVA 1.5 & 52.12 & 52.29 & 59.27 & 58.99 & 63.63 & 50.67 & 49.70 & 48.96 & 48.56 & 63.66 & 38.99 & 38.37 & 52.29 \\
        LLaVA1.6 & 57.24 & 68.31 & 61.22 & 69.02 & 63.35 & 64.62 & 65.07 & 66.52 & 47.69 & 45.29 & 43.95 & 47.46 & 44.40 \\
        Qwen2.5-VL & 47.26 & 41.81 & 44.52 & 44.60 & 48.45 & 48.70 & 45.57 & 43.01 & 56.62 & 44.69 & 52.18 & 49.13 & 47.88 \\
        Qwen3-VL & 43.73 & 48.25 & 52.09 & 48.45 & 53.29 & 54.06 & 46.08 & 51.30 & 45.51 & 28.45 & 35.86 & 33.52 & 27.88 \\
        \midrule
        Random & 50.00 & 50.00 & 50.00 & 50.00 & 50.00 & 50.00 & 50.00 & 50.00 & 50.00 & 50.00 & 50.00 & 50.00 & 50.00 \\
        Human Estimate & 98.17 & 96.00 & 98.00 & 98.00 & 98.00 & 98.00 & 100.00 & 100.00 & 98.00 & 96.00 & 100.00 & 100.00 & 96.00 \\
        \bottomrule
    \end{tabular}
    }
    \caption{All models perform poorly on the VCR-Causal benchmark, only slightly better than random guess. CW1--CW12 refer to the same twelve causal conjunctions defined in Table~\ref{tab:vqacausalresults}. ``BLIP2 FE'' denotes the BLIP2 feature extractor model.}
    \label{tab:vcrcausalresults}
    \vspace{-1em}
\end{table*}

\mypar{VQA-Causal Construction} We leverage GPT-4-turbo to generate the captions for our VQA-Causal and VCR-Causal benchmarks. For each image in our benchmarks, we extract its question and set of answers from the original annotation files, and then feed both the question and answers into GPT-4-turbo to produce two causal sentences. Concretely, we use the following prompt:

\begin{PromptBox}
Here is a question and several possible answers. Summarize the most reasonable answer and create a sentence combining the question and the answer using "is due to" or "is caused by". Please ensure that both parts of the sentence before and after "is due to" or "is caused by" use specific subjects rather than pronouns like it/he/she/them. Then, generate a second sentence by reversing the causal order of the first sentence and using "is due to" or "is caused by" regardless of whether the sentence makes logical sense. For example, if the first sentence is "There is water on the ground is due to it rained," the second sentence would be "It rained is due to there is water on the ground." Ensure that you return exactly 2 sentences as per my request, and present the two sentences on separate lines, nothing else. \\
\smallskip
\noindent For example:
        There is water on the ground is due to it rained.
        It rained is due to there is water on the ground. 

Question: “\{question\}”  
Answers: \{result\}
\end{PromptBox}

\mypar{Activity and Object Understanding Test} We leverage GPT-4-turbo to generate extended captions for the images in our Activity and Object Understanding Test. For each image, we supply two existing captions describing the image to GPT-4-turbo, and request expanded variations. Specifically, we use the following prompt:

\begin{PromptBox}
Based on the sentence: \{sentence\}, generate a new sentence that includes the object category of the subject from the original sentence, but with different content or a new sentence that includes the same activity but with different object. For example, you can use the object category of the subject from the original sentence as the object in the new sentence, introduce a new object as the subject, and also change the verb, color, weather, gender, and other elements to be different from those in the original sentence. Return only the newly generated sentence; nothing else.
\end{PromptBox}

\end{document}